\documentclass[10pt,letterpaper]{article}

\usepackage{cogsci}

\cogscifinalcopy 

\usepackage{pslatex}
\usepackage{apacite}
\usepackage{float} 

\usepackage[hyphens]{url} 
\usepackage{graphicx}

\title{Bridging the prosody GAP: Genetic Algorithm with People to efficiently sample emotional prosody}

\author{{\large \bf{Pol van Rijn}\textsuperscript{1,*}, Harin Lee\textsuperscript{1,2,*} and Nori Jacoby\textsuperscript{1}} \\
  \\
  \textsuperscript{1}Max Planck Institute for Empirical Aesthetics, Frankfurt am Main, Germany \\
  \textsuperscript{2}Max Planck Institute for Human Cognitive and Brain Sciences, Leipzig, Germany \\
  \\
  \textsuperscript{*}Equal Contribution, Corresponding Authors: \{pol.van-rijn, harin.lee\}@ae.mpg.de
  }

\begin{document}

\maketitle

\begin{abstract}

The human voice effectively communicates a range of emotions with nuanced variations in acoustics. Existing emotional speech corpora are limited in that they are either (a) highly curated to induce specific emotions with predefined categories that may not capture the full extent of emotional experiences, or (b) entangled in their semantic and prosodic cues, limiting the ability to study these cues separately. To overcome this challenge, we propose a new approach called `Genetic Algorithm with People’ (GAP), which integrates human decision and production into a genetic algorithm. In our design, we allow creators and raters to jointly optimize the emotional prosody over generations. We demonstrate that GAP can efficiently sample from the emotional speech space and capture a broad range of emotions, and show comparable results to state-of-the-art emotional speech corpora. GAP is language-independent and supports large crowd-sourcing, thus can support future large-scale cross-cultural research.

\end{abstract}

\section{Introduction}

The human voice contains rich paralinguistic information that can alter or give nuanced meaning to what we say and communicate how we feel \cite{banse1996}. This paralinguistic information is transmitted through speech prosody, which refers to the variations in pitch, loudness, timing, and voice quality. However, how emotions are exactly communicated with voice is not fully understood. The most prominent way of studying the mapping between emotions and speech prosody has been to investigate the relationship between acoustic features and emotion labels in given large corpora of spoken sentences \cite{elayadi2011}.


Existing emotional speech corpora are mainly constructed in two ways: (1) corpora in which annotators label recordings by their perceived emotion, and (2) corpora in which participants are prompted to express certain emotions in their speech. In the first case (perceived emotion), the recordings often consist of naturalistic speech, either scraped from naturalistic contexts (e.g., YouTube clips, TV shows, debates), or recorded from participants who are engaged in game-like activities (e.g., acting improvisation, \citeA{IEMOCAP}, user interactions with robots, \citeA{FAU-AIBO}). Given emotional events rarely occur in naturalistic settings, these corpora are often limited in the number of annotations and time-consuming to collect. Furthermore, emotional information is encoded simultaneously through multiple channels \cite{paulmann2011} but some channels are entangled with one another and cannot easily be separated. For instance, \textit{what} we say and \textit{how} we say both influence the listener's perception of emotions. Such interaction is problematic because we cannot dissect whether the judgment provided by the annotators arises from the semantic meaning, prosody, or combination of the two. On the other hand, intended emotion corpora are often experimentally well-controlled, balanced, and contain high-quality audio recordings. However, they are limited to predefined categories of emotions, lack naturalness, and are often expensive to create -- thus compromising by reducing the number of recordings.

\begin{figure*}[ht!]
    \centering
    \includegraphics[width=\textwidth]{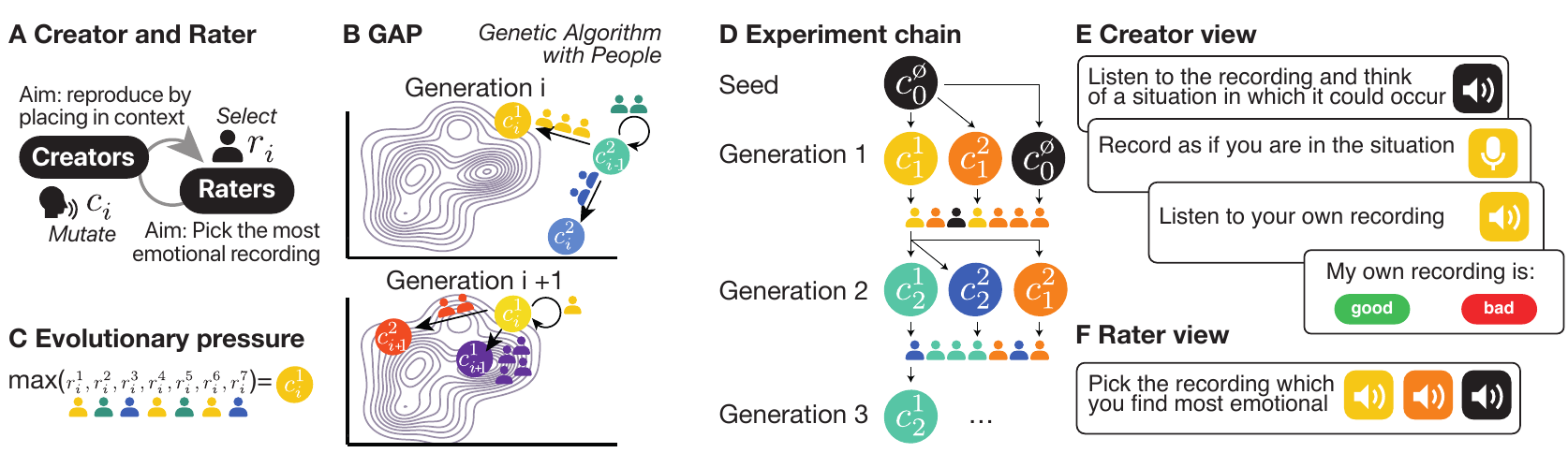}
    \caption{Schematic diagram of the main experiment. (A) Creators generate mutant recordings of the previous generation by recording themselves saying the sentence as if they were in the same situation. Raters select the most emotional recording and hence the Darwinian selection is applied. (B) The raters are presented with $n$ creations, whereby $n-1$ stimuli are created in the current generation $i$. Over the generations, the speech prosody moves closer towards the prototype of a particular emotion. (C) Raters put on a majority vote to decide which recording should propagate to the next generation. (D) Each chain starts with an initial creation $c_0^{\o}$, which is passed on to the first creators $c_1$. The raters rate the creations $\{c_{i-1}, c^1_i, \dots, c^m_i\}$ where $m$ is the number of creators (here $m = 2$). (E) Each creator first listens to the creation from the previous generation and has to think of a situation in which the recording could occur. Next, they record themselves as if they were in the same situation as the previous speaker, followed by a playback of their recording to confirm that the recording is correct. (F) Raters listen to $m + 1$ recordings (random order) and select the most emotional recording.}
    \label{fig:methods}
\end{figure*}

Here, we introduce a new method to construct corpora for emotional speech. Our design extends from previous methods that integrate human decisions into optimization algorithms, such as MCMCP \cite{sanborn2008} and GSP \cite{harrison2020a}. We propose to combine human production and decision with a genetic algorithm to efficiently search the high-dimensional space containing all possible articulations of emotional speech. Genetic algorithms typically include two steps: mutation and selection (Figure \ref{fig:methods}A). Our proposed method, called `Genetic Algorithm with People' (GAP), implements mutation and selection by assigning roles to people as \textit{creators} and \textit{raters}. The creator's role is to hear the spoken sentence from the previous generation and reproduce it with their recording by situating themselves in the context of the heard speaker. The raters then put on a majority vote among three recordings (mutants) and decide which recording should be propagated to the next generation (i.e., Darwinian selection, Figure \ref{fig:methods}B). In our design, we ask the raters to select the most emotional recording, thus increasing the evolutionary pressure of emotional recordings and boosting the occurrence of possible emotional events. Since creators are not prompted to produce emotional speech, they are less likely to produce emotional stereotypes. Furthermore, we can disentangle semantic content from prosody because the same spoken sentence propagates throughout the generations, allowing us to control for the influence of semantic cues on emotion perception. 

We argue that our proposed method -- Genetic Algorithm with People (GAP) -- is less-biased because it does not presume the dimensionality of the emotional space and is more efficient in sampling emotional prosody. Moreover, because the recordings can be collected online, it can considerably reduce the expenses and resources that are often necessary for a more traditional corpora curation (inviting participants to the lab, booking recording studios, manual annotation, etc). GAP is also scalable in size by benefiting from the large and diverse pool of online participants and can easily be extended to other languages to produce a multi-lingual corpus. Although we only focus on the English language in the current study, the method is not limited to and can be applied to any language given there are no prerequisites (e.g., pre-trained language embedding, language-specific pre-screener). Finally, GAP can be applied to other modalities, in which different kinds of evolutionary pressure can be introduced. For example, optimizing the simplicity of a hand-drawn object \cite{schaldenbrand2021styleclipdraw, tian2022modern} or the aesthetics of music \cite{maccallum_evolution_2012}.\footnote{All stimuli generated in the experiment chains can be explored though an online, interactive visualization: \url{https://polvanrijn.github.io/prosody-GAP/}}

\section{Methods}
Both experiments reported in this paper were conducted using PsyNet (\url{https://PsyNet.dev}; \citeauthor{harrison2020a}, 2020), an automatic online recruitment system that is based on the Dallinger platform (\url{https://github.com/Dallinger/Dallinger}). All participants provided informed consent in accordance with the Max Planck Society Ethics Council approved protocol (application 2020\_05).

\subsection{Experiment 1: Genetic Algorithm with People}
\paragraph{Stimulus generation}
Prior to conducting the experiment, we compiled a list of 10 semantically neutral sentences (see Supplementary Materials). Eight of these sentences are included in the two emotional corpora, which are later used in the validation experiment \cite{CREMA-D, VENEC}. The remaining two sentences come from the phonetically balanced and semantically neutral Harvard sentence corpus \cite{harvardsentences}. For each sentence, we generated five speech recordings each with a different speaker using the expressive TTS model \cite{flowtron} trained on LibriTTS (2,456 speakers). We manually inspected if recordings contained glitches and replaced them with new ones when necessary.

\paragraph{Design}
We employed a chain experiment design, whereby the artificially generated samples served as the initial seed generation. Participants were divided into two groups: \textit{creators} who reproduced the speech, and \textit{raters} who rated the recordings generated by the creators. Each chain consisted of 10 generations (including the initial generation), totaling 500 recordings produced by the creators and judged by the raters.

\paragraph{Procedure}
In each chain, the creators first heard the recordings of the previous generation and were asked to situate themselves in the context of the speaker and repeat the sentence. After completing the recording, they could hear the playback of their own recording and evaluate whether their own recording was good or bad (Figure \ref{fig:methods}E). Only when the creator confirmed that his or her recording was of good quality, their own recording could be subsequently rated by the rater group. For the creators, we explicitly chose to avoid the use of words such as ``emotions" or ``feelings" in the experiment text and instead instructed the creators to place themselves in the context of the speaker. This was to minimize the potential biases in prompting participants to produce stereotypical emotions.

The rater group heard three alternative recordings, of which two were recordings produced by the creators in the current generation (i.e., mutants) and the other one being the selected recording of the previous generation. The raters were asked to select which of the recording was most emotional (Figure \ref{fig:methods}F). In each generation, seven rater responses were gathered and the recording with the majority vote was propagated to serve as the stimulus for the next generation (Figure \ref{fig:methods}C). Consistent with previous literature, in pilot experiments, we found that introducing a majority voting  approach reduces participant error \cite{krishna_visual_2017} and improves the quality of productions \cite{maccallum_evolution_2012}. Each participant had a fixed role because raters and creators have different tasks that could potentially influence one another.

\subsection{Experiment 2: Annotation}
To validate the robustness of our paradigm, we recruited an independent group of participants to provide annotations for the recordings generated in Experiment 1, as well as the recordings obtained from two existing emotional prosody corpora. We compared the three datasets to examine their quality and breadth of coverage of the emotion prosody space.

\paragraph{Materials}
Among the 10 neutral sentences we chose for Experiment 1, we subset 20 recordings each from the 6 categories of emotion in CREMA-D \cite{CREMA-D}, and 10 recordings each from the 11 categories of emotions in the US subset of VENEC, totaling 230 stimuli. These two high-quality corpora obtained in a traditional way were added as a baseline for comparison with GAP. From both corpora, we created another subset of 30 recordings that were intended to be neutral.

The 500 recordings obtained from the genetic reproduction of prosody had duplicates since some chains converged early (i.e., the previous generation stimulus being selected again because it was most emotional). Removing these duplicates resulted in 314 unique recording to compose the genetic prosody set (hereafter referred as `prosody-GAP'). The final stimuli set of all three datasets combined summed to 544 recordings.

\paragraph{Design}
We employed a within-participant design, in which each participant was presented with 20 randomly drawn recordings from the entire stimulus set. Additionally, we presented 2 extra repeated trials to measure the consistency of their response which was later used for exclusion criteria.

\paragraph{Procedure}
For each recording, participants answered questions presented in the following order: perceived strength of emotion (``how emotional was the speech?", 4-point scale), valence (``how negative or positive was the speech?", slider value ranging -50 to +50) and arousal (``how low or high in energy was the speech?", slider value ranging 0 to +100), and the authenticity of the speech (``how authentic (real) or fake (pretending) was the speech?", 4-point scale). Additionally, they were asked to type a single word related to the mood that best describes the state of the speaker in the recording.


\subsection{Participants}
991 participants were recruited from Amazon Mechanical Turk (MTurk) who satisfied the criteria of having a US residency with a lifetime approval rate of over 99\% on the platform. All participants had to pass a series of screening tasks before they could participate in the main experiment. These screening tasks included an assessment of competence in English \cite{lextale} and a psychoacoustic test to check whether they were using headphones \cite{woods2017headphone}.

To obtain better quality audio recordings, we applied stricter screening criteria in Experiment 1 where the participants had to have more than 2,000 previously completed tasks on MTurk. Creators additionally had to pass the two screening tasks: first, they had to be able to distinguish good from bad recordings. Bad recordings were recordings that were silent, contained too much noise, repeated sentences, or sentences cut out too early. If participants made more than one mistake, they were excluded. Second, participants were asked to reproduce the heard sentences during practice trials and were excluded if there was a textual mismatch (identified using Google's speech-to-text API transcription).

An additional screening criterion was also applied in Experiment 2. We excluded participants who repeatedly gave the same answers for text labels and ones who had low response consistencies ({$r < $}.40) between the main experiment trials and the repeated trials at the end.

After the screening procedure, 126 participants 
remained for Experiment 1 with a mean age of 39.0 (SD = 10.3), among which 28 had the role of being creators and 98 being raters. 131 participants remained for Experiment 2 with a mean age of 38.3 (SD = 11.9). All participants received monetary compensation at a rate of \$9 an hour for their participation.

\begin{figure*}[ht!]
    \centering
    \includegraphics[width=\textwidth]{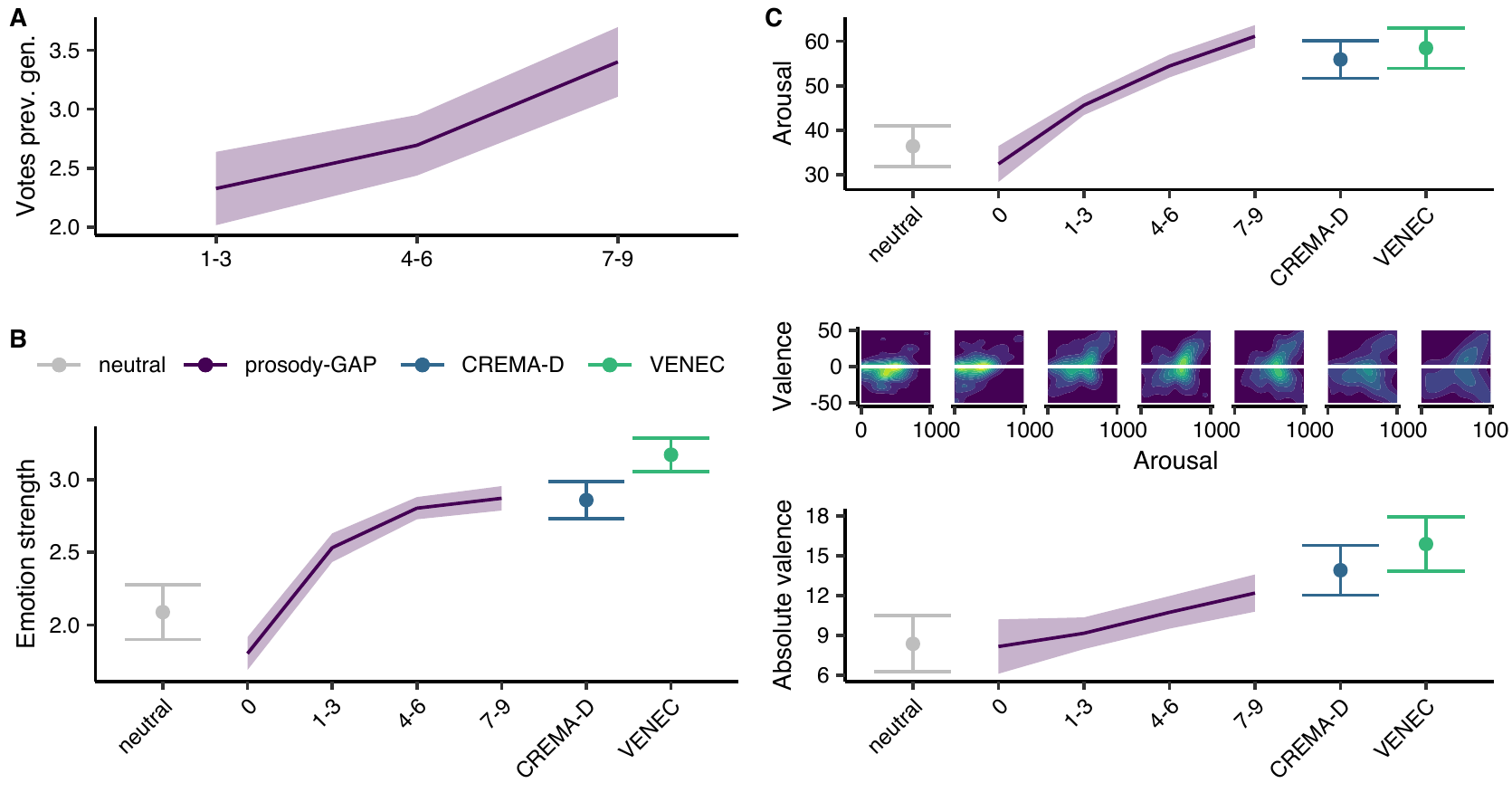}
    \caption{Emotion validation. (A) Average number of raters selecting the stimulus from the previous generation with the generations binned into 1--3, 4--6, and 7--9. The number of raters range from 0 to 7, in which 7 would mean that all raters selected the creation from the previous generation. (B) Average rating on the strength of emotions for the neutral stimuli (gray) and emotional stimuli from the VENEC (blue) and CREMA-D (green) corpus. The initial generation of prosody-GAP as bin 0 and the following generations binned into 1--3, 4--6, and 7--9. The area and the error bars represent 95 \% confidence intervals. (C) Average arousal (top) and absolute valance (bottom) as well as the arousal and valence kernel density estimates (middle).}
    \label{fig:emotion-validation}
\end{figure*}

\section{Results}
\paragraph{Natural Emergence of Emotional Speech}
First, we analyzed the selections made by the rater groups in Experiment 1, where the recording with the majority vote survived to the next generation. If it were that the speech recordings became more emotional in each generation, the raters would choose one of the new recordings generated in the current generation as opposed to choosing the one from the previous generation. In line with this, over the generations, we could observe that the raters chose the new mutant recording to be more emotional than the one from the previous generation (see Figure \ref{fig:emotion-validation}A). 

Second, we analyzed the strength of emotions in the recording and their valence and arousal response obtained from the annotators of Experiment 2. We then compared this with the existing emotional prosody corpora. In-line with our prediction, the seed stimuli (generation 0) of the artificially generated recordings with GAP showed similar ratings ({$M$} = 1.79, {$SD$} = 0.80) as the neutral stimuli of CREMA-D and VENEC combined ({$M$} = 2.09, {$SD$} = 0.86; see Figure \ref{fig:emotion-validation}B). Furthermore, the kernel density estimation (KDE) showed dense concentration around the center of the valence and arousal 2-dimensional space (see Figure \ref{fig:emotion-validation}C). The seed of prosody-GAP and the neutral sets of the other two corpora also showed comparable levels of arousal and absolute valence (after first averaging at the stimulus level with raw values).

Over the generations of prosody-GAP, we observed that the recordings became gradually more emotional and reached a plateau around the 6th generation (see Figure \ref{fig:emotion-validation}B), of which the rating of the last generation ({$M$} = 2.79, {$SD$} = 0.86) were slightly higher than CREMA-D ({$M$} = 2.70, {$SD$} = 0.97) but lower than VENEC ({$M$} = 3.11, {$SD$} = 0.88). Moreover, the coverage of the valence-arousal space dispersed over the generation, and by the last generation, we could observe that the covered regions were similar to CREMA-D and VENEC -- judging by visual inspection of the KDE plot (see Figure \ref{fig:emotion-validation}C).

Overall, these results demonstrate the robustness of GAP and suggest that (i) emotional speech can be obtained in a less biased way without the prior assumptions of emotion categories, and (ii) the obtained recordings from online crowd-sourced samples can achieve comparable results to carefully curated corpora generated in professional settings. Furthermore, the convergence of emotional levels around the 6th generation demonstrates the efficiency of our method and shows highly promising potential for its scalability.

\begin{figure*}[ht!]
    \centering
    \includegraphics[width=\textwidth]{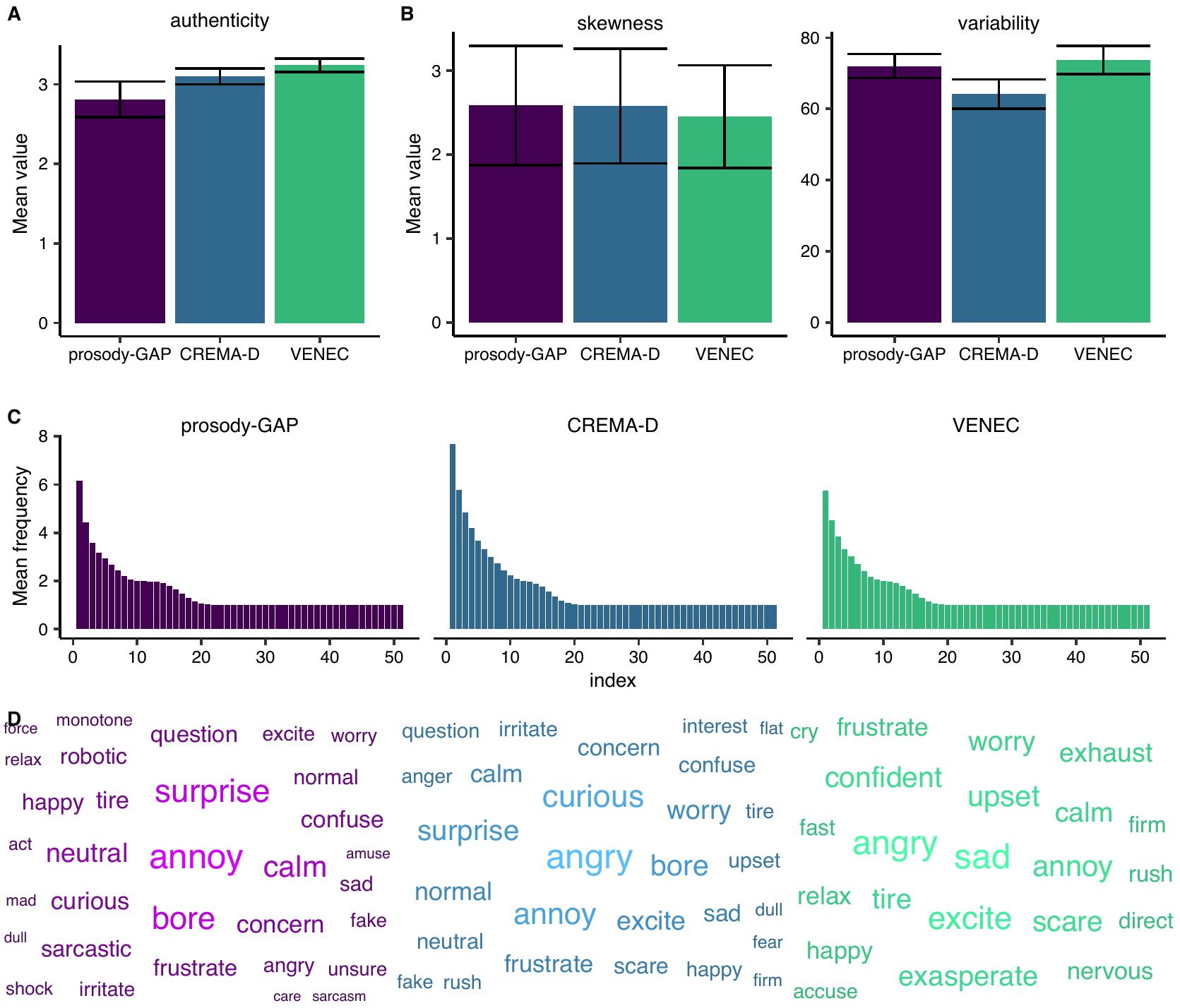}
    \caption{Authenticity validation and free-text responses. (A) Distribution of authenticity ratings for stimuli from our corpus prosody-GAP, VENEC, and CREMA-D. The error bars correspond to 95 \% confidence intervals. (B) Computed skewness of the frequency distribution of the free-text response (1,000 bootstraps). The error bars correspond to standard deviations. (C) The average sampled word frequency distributions with bootstrap sample size being the smallest distribution length to balance size across the three datasets. (D) Wordcloud generated from the free text response.}
    \label{fig:authenticity-labels}
\end{figure*}

\paragraph{Samples from a Wide Array of Emotions}
Using the gathered word labels provided by the annotators in Experiment 2, we quantified the variability and the term-frequency distributions and made comparisons across the three datasets.

For prosody-GAP, we took the stimulus of the last generation where the ratings converged in each of the 50 independent chains (i.e., recording that the rater group judged as most emotional). Since the size of the stimuli were unbalanced across the three sets (prosody-GAP = 50, CREMA-D = 100, VENEC = 100; excluding neutral stimuli), we sampled 50 stimuli from CREMA-D and VENEC at random to match the stimuli size. We then computed 1,000 bootstraps without replacement in each set by randomly drawing 100 word label samples from all responses (where each stimulus can have multiple annotations made by independent annotators). All word labels were lemmatized using the \texttt{textstem} R package.

To measure variability, we counted the number of unique word labels in each of the bootstrapped samples (see Figure \ref{fig:authenticity-labels}A). The results showed that both prosody-GAP ({$M$} = 72.1, {$SD$} = 3.30) and VENEC ({$M$} = 73.9, {$SD$} = 3.85) obtain comparable variability, and higher value on average than CREMA-D ({$M$} = 64.0, {$SD$} = 4.16). High variability indicates that more diverse semantic labels are present, covering a wider range of semantic vocabulary associated with emotions, whereas low variability suggests that the frequency of words is concentrated onto a smaller subset of words. Considering VENEC consists of more emotion categories than CREMA-D (11 and 6, respectively), higher variability for VENEC was expected. The fact that prosody-GAP achieves comparable variability to VENEC is indicative of the large breadth of emotion space GAP is able to capture.

We next explored the raw distributions of the bootstrapped samples. Since each bootstrapped sample has a varying number of entries, we first balanced this by taking the smallest number of unique word values among all bootstrapped samples and used this as the reference value to consider the distributions only over this threshold (see Figure \ref{fig:authenticity-labels}C); therefore allowing us to focus on the most informative portion of the distribution. We then aggregated over the bootstrapped samples in each set and computed the skewness to investigate the distribution of the words, whereby high skewness in distribution would indicate that a small subset of words occurs more frequently and that there is more inequality. Across all three datasets, the skewness was highly similar (see Figure 3B) and the top word labels in each set are visualized as wordclouds in Figure 3D.

\paragraph{Authenticity}
In Experiment 2, annotators made a judgment about how authentic (real) or fake (pretending) the recordings were (see Figure 3A). One-Way between-group ANOVA showed that there was significant group differences ({$F(2, 390)$} = 12.22, {$p <$} .001, {$ges$} = .06) with prosody-GAP being rated as least authentic ({$M$} = 2.87, {$SD$} = 0.52) while CREMA-D ({$M$} = 3.03, {$SD$} = 0.58) and VENEC ({$M$} = 3.21, {$SD$} = 0.58) obtaining similar ratings. However, the effect size between the groups was small, and a large portion of the stimulus in all three sets were rated as either “somewhat authentic” or “very authentic”. Given prosody-GAP was produced in a noisier and uncontrolled environment recorded by non-voice actors, we interpret this as an encouraging result that corpora of sufficient quality can be obtained using online participants.

\section{General Discussion}
We propose a new approach, Genetic Algorithm with People (GAP), for efficient sampling of the high-dimensional emotion prosody space by introducing genetic algorithms with human raters. Each participant is assigned the role of either creator or rater and the two groups synchronously optimize the emotion prosody over each generation (Experiment 1). While existing emotional prosody corpora are generated in a top-down fashion relying on predefined emotion categories, we show that our method can achieve comparable results to produce corpora using online participants and by allowing the emotions in speech to naturally emerge.

Using GAP, the speech became more emotional over the generations and naturally converged, capturing a wide range of possible emotions (Figure \ref{fig:authenticity-labels}). This was assessed by an independent group of annotators that provided responses for perceived strength of emotion, valence and arousal (Experiment 2). In the beginning, we observed that our seed stimuli (i.e., generation 0) had similar ratings and KDE concentration to the neutral stimuli of CREMA-D and VENEC (Figure \ref{fig:emotion-validation}). As each successive generation optimized to maximize the emotions in the recording through selection pressure, the level of emotion, valence, and arousal gradually inclined to reach a plateau at a value that was comparable with the other corpora. At the same time, it showed similar dispersion in the KDE plot.

When examining the term-frequency distributions and the variability of provided word labels to speech recordings, we observed that the distributions among the three datasets were similar and that the variability of words obtained using GAP was comparable to VENEC and better than CREMA-D. These results highlight the robustness of GAP and its great potential for generating emotional prosody corpora in a curated environment that does not require professionals, with no a priori assumptions about the dimensionality of the emotional space. This makes it a particularly valuable tool for conducting research on emotional prosody cross-culturally and for low-resource languages in particular. Recent findings have shown that the use of emotion semantics in text are variable across cultures \cite{thompson_cultural_2020} and that perception of emotions in music may be influenced by listener's cultural background \cite{lee_cross-cultural_2021}. Thus, GAP is a great alternative to overcome such challenges as it does not rely on assumptions of emotion categories and can be extended to any language.

Our design is still limited in that we restrict the annotators to provide only a single label for the recording (when there could be a range of emotions perceived) and some recordings had -- in spite of extensive screening tasks -- poor audio quality. Also, it is possible that maximizing the emotion in the selection task biases the outcome towards prototypical emotions. However, the prior is likely to be less biased than existing corpora that rely on inducing specific and predefined emotions. In future work, we plan to remove constraints on the number of labels, develop a dynamic convergence metric to end chains earlier, and construct a more principled way for selecting sentences for initial generations that do not have to rely on an existing corpus. Furthermore, we wish to implement better audio control to screen for participants with bad microphone quality to improve the overall recording quality of the corpora. 

GAP has much broader applications and can be extended to other domains for optimizing stimuli to match subjective human criteria. For example, the method could be applied in the context of optimizing sung melodies or hand drawings to find optimal aesthetic preferences and compare across different groups. More broadly, this method demonstrates the application of integrating humans in computer optimization techniques to solve long-standing problems in machine learning and perception.


\bibliographystyle{apacite}

\setlength{\bibleftmargin}{.125in}
\setlength{\bibindent}{-\bibleftmargin}

\bibliography{CogSci_Template}

\end{document}